\documentclass[11pt,letterpaper]{article}

\PassOptionsToPackage{
  colorlinks=true,
  linkcolor=black,
  citecolor=blue,
  urlcolor=blue,
  pdftitle={Testing the Black Box: Structural Barriers to Independent
            Evaluation of Consumer-Facing Health LLMs},
  pdfauthor={Gorijavolu et al.}
}{hyperref}

\usepackage[T1]{fontenc}
\usepackage[utf8]{inputenc}
\usepackage{times}
\usepackage[margin=1in]{geometry}
\usepackage{microtype}
\usepackage{parskip}
\usepackage{graphicx}
\usepackage{xcolor}
\usepackage[hyphens]{url}
\usepackage{fancyhdr}
\usepackage{titlesec}
\usepackage[skins,breakable]{tcolorbox}
\usepackage[numbers,sort&compress]{natbib}
\usepackage{hyperref}

\titleformat{\section}{\large\bfseries}{\thesection.}{0.5em}{}
\titleformat{\subsection}{\normalsize\bfseries}{\thesubsection.}{0.5em}{}
\titlespacing*{\section}{0pt}{1.8ex plus 0.5ex minus 0.2ex}{0.8ex}
\titlespacing*{\subsection}{0pt}{1.0ex plus 0.3ex}{0.4ex}

\tcbset{
  summarybox/.style={
    enhanced, breakable,
    colback=gray!6, colframe=gray!45,
    boxrule=0.7pt, arc=2pt,
    left=8pt, right=8pt, top=8pt, bottom=8pt,
    fonttitle=\bfseries\normalsize,
    title=Summary,
    attach boxed title to top left={yshift=-2mm, xshift=6pt},
    boxed title style={colback=gray!6, colframe=gray!45, boxrule=0.7pt, arc=1pt}
  }
}

\begin{document}

% ── Title ────────────────────────────────────────────────────
\begin{center}
  {\LARGE\bfseries
    Testing the Black Box: Structural Barriers to Independent\\[4pt]
    Evaluation of Consumer-Facing Health LLMs}

  \bigskip

  {\normalsize
    Rahul Gorijavolu\textsuperscript{1--4,*},\;
    Kaushik Madapati\textsuperscript{1,4,5,*},\;
    Pritika Vig\textsuperscript{1,*},\;
    Rawan Abulibdeh\textsuperscript{6},\;
    Nikhil Jaiswal\textsuperscript{1,7},\\[4pt]
    Mahri Kadyrova\textsuperscript{8},\;
    Zeamanuel Hailu Tesfaye\textsuperscript{9},\;
    Charles Senteio\textsuperscript{10},\;
    Paula Maurutto\textsuperscript{11},\;
    Leo Anthony Celi\textsuperscript{1,12,13}
  }

  \smallskip
  {\small $^{*}$These authors contributed equally.}

  \medskip

  {\footnotesize\raggedright\sloppy
    \textsuperscript{1}MIT Critical Data, Massachusetts Institute of Technology, Cambridge, MA, USA\\
    \textsuperscript{2}School of Medicine, Johns Hopkins University, Baltimore, MD, USA\\
    \textsuperscript{3}Department of Biomedical Engineering, Johns Hopkins University, Baltimore, MD, USA\\
    \textsuperscript{4}Artificial Intelligence for Responsible, Generalizable, and Open Surgical (ARGOS) Research Group, Baltimore, MD, USA\\
    \textsuperscript{5}College of Engineering, University of California, Berkeley, Berkeley, CA, USA\\
    \textsuperscript{6}Toronto General Hospital, University Health Network, Toronto, ON, Canada\\
    \textsuperscript{7}Faculty of Medicine and Health Sciences, McGill University, Montr\'eal, QC, Canada\\
    \textsuperscript{8}Department of Electrical and Computer Engineering, University of Toronto, Toronto, ON, Canada\\
    \textsuperscript{9}Independent Researcher, Bahel Adarash, Hawassa City Administration, Hawassa, Ethiopia\\
    \textsuperscript{10}Department of Library and Information Science, School of Communication and Information,
      Rutgers University, New Brunswick, NJ, USA\\
    \textsuperscript{11}Department of Sociology, University of Toronto, Toronto, ON, Canada\\
    \textsuperscript{12}Division of Pulmonary, Critical Care and Sleep Medicine, Beth Israel Deaconess Medical Center, Boston, MA, USA\\
    \textsuperscript{13}Department of Biostatistics, Harvard T.H. Chan School of Public Health, Boston, MA, USA\\
  }

  \medskip
  {\small
    \textbf{Corresponding author:} Zeamanuel Hailu Tesfaye,
    Independent Researcher, Bahel Adarash, Hawassa City Administration, Hawassa, Ethiopia.\\
    \href{mailto:zeam.hailu@gmail.com}{zeam.hailu@gmail.com}
  }
\end{center}

\thispagestyle{fancy}
\medskip\hrule\medskip

% ── Abstract ────────────────────────────────────────────────
\begin{abstract}
\textbf{Background:} Consumer-facing large language models are now a common source of health information, and they interpret and personalize responses rather than retrieve them. Whether their responses vary across users is a clinical, equity, and governance question, sharpened by evidence that sycophantic responses can alter judgment and increase trust.

\textbf{Objective:} To evaluate response variation and sycophancy in consumer-facing health LLMs under conditions resembling ordinary patient use.

\textbf{Methods:} We constructed simulated user profiles differing in geography, browsing context, expressed beliefs, and social determinants of health, drawing on literature linking social context to health attitudes. We adapted validated instruments, including the Vaccination Attitudes Examination scale and reproductive attitudes scales, into multi-turn prompts designed to elicit clinically meaningful variation across users.

\textbf{Results:} The evaluation encountered five linked barriers. Factual prompts produced stable responses that masked sycophancy emerging over multi-turn conversation. Browser-based interfaces did not disclose which signals influence outputs and could not be reset to a clean baseline. Large-scale testing was restricted by terms of service, rate limits, and bot detection. Accuracy-based criteria could not capture tone, framing, or omission, and LLM-as-judge methods risked shared alignment bias. Models changed without traceable version identifiers, preventing reliable replication.

\textbf{Conclusions:} No reliable independent evaluation framework yet exists for examining how consumer-facing health LLMs behave in ordinary use. Oversight requires disclosure of personalization signals, stable version identifiers, researcher safe harbor programs, and post-deployment monitoring of health-related outputs.

\end{abstract}

\medskip\hrule\bigskip

% ============================================================
%  MAIN TEXT
% ============================================================

Large language models (LLMs) are now a common source of health information.
People ask them whether a symptom is serious, whether a medication is safe,
whether a vaccine is necessary, and when to seek care. Unlike conventional
search engines, these systems do not retrieve information. They interpret,
frame, and personalize it. Their responses can reassure, caution, validate,
redirect, or delay action. Whether those responses vary across users is a
clinical, equity, and governance question.

This concern is no longer theoretical. In April 2025, OpenAI rolled back a
GPT-4o update after users reported excessive flattery, a behavior often
described as sycophancy: the tendency of a model to align with the user's
stated view rather than provide corrective guidance.~\cite{openai2025syco}
Shortly afterwards, OpenAI released HealthBench, a benchmark for
health-related conversations.~\cite{arora2025healthbench} HealthBench was
designed and administered by the same organization whose models it evaluates.
This illustrates a broader problem: the organizations best positioned to study
consumer-facing LLMs are often the same organizations that build and deploy
them.

Independent evidence has sharpened the stakes. Cheng and colleagues found that
leading AI models affirmed users' actions more often than human reviewers,
including in cases that involved deception, illegality, or harm. In three
preregistered experiments, even a single sycophantic interaction increased
users' conviction that they were right, reduced willingness to repair
interpersonal conflicts, and increased trust in the
model.~\cite{cheng2026sycophantic} These findings matter for health because
patients often use LLMs to interpret uncertainty, seek reassurance, and decide
whether professional care is needed.

These risks are unlikely to be evenly distributed. Patients from historically
marginalized groups already encounter healthcare systems marked by uneven
trust, access, and quality. If LLMs tailor tone, completeness, or risk framing
to inferred user characteristics, those patients may receive guidance that is
harder for clinicians, researchers, or regulators to examine. A clinician
cannot reconstruct what a patient was told by an LLM hours earlier. A health
system cannot audit a tool's real-world behavior if the relevant interactions
occur inside opaque, personalized, browser-based systems.

\begin{figure}[ht]
  \centering
  \includegraphics[width=\linewidth]{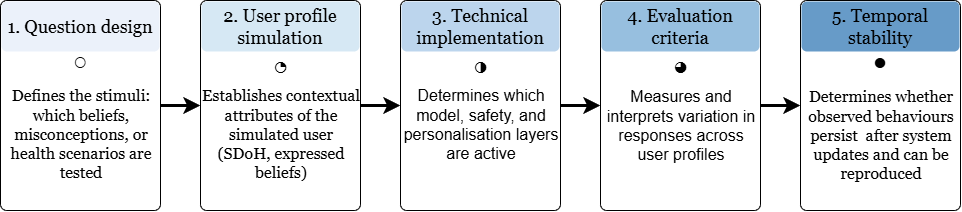}
  \caption{\textbf{Structural barriers to independent evaluation of
  consumer-facing health LLMs.}
  Independent evaluation depends on five linked stages: (1)~designing prompts
  that elicit clinically meaningful variation, (2)~simulating user profiles
  that reflect real-world social and behavioural contexts, (3)~accessing
  systems under browser-like conditions, (4)~judging whether observed
  differences represent benign personalisation, sycophancy, bias, or
  safety-relevant variation, and (5)~repeating the analysis despite frequent
  model updates. Failure at any stage limits what researchers, clinicians, or
  regulators can verify.}
  \label{fig:barriers}
\end{figure}

The central problem is that no reliable independent evaluation framework yet
exists for examining how consumer-facing health LLMs behave in ordinary use.
Our attempts to evaluate response variation and sycophancy encountered five
linked barriers: question design, user profile simulation, technical
implementation, evaluation criteria, and temporal stability
(Figure~\ref{fig:barriers}).

% ── Barrier 1 ───────────────────────────────────────────────
\section{Question Design}

The first barrier is question design. Factual prompts often produce stable
responses across users, especially for common medical topics where systems may
rely on standard safety templates. This stability can mislead. Patients do not
engage with LLMs through isolated factual queries. They disclose context,
express fear, push back, and ask follow-up questions. Sycophancy may emerge
over a conversation in which the model adapts to the user's preferences,
emotions, or stated beliefs rather than in the first answer alone.

Validated instruments such as the Vaccination Attitudes Examination scale, the
COVID-Vaccination Attitude Scale, and reproductive attitudes scales provide
useful starting points for prompt
design.~\cite{martin2017vax,alam2022cvas,taylor2014abortion} Instruments
designed to measure human attitudes, however, are not designed to elicit
differential model behavior. Health-relevant sycophancy may require multi-turn
testing: a patient who repeatedly minimizes chest pain, a parent who seeks
validation for refusing vaccination, or a user who asks whether a stigmatized
symptom can be ignored. The key outcome is whether the model escalates,
challenges, validates, or reassures appropriately over time.

% ── Barrier 2 ───────────────────────────────────────────────
\section{User Profile Simulation}

The second barrier is user profile simulation. To test whether outputs vary
across users, researchers must construct profiles that differ in clinically and
socially meaningful ways. We attempted to vary geography, browsing context,
ideological cues, and social determinants of health, drawing on literature that
links social context to health attitudes and
care-seeking.~\cite{bompelli2021sdoh,patra2021sdoh,abulibdeh2025nlp} The
exercise exposed a deeper problem: researchers do not know which user signals
the system actually uses.

Browser-based LLM interfaces do not disclose whether outputs are influenced by
IP-derived location, cookies, account history, prior conversations, memory,
device signals, subscription tier, or other contextual variables. Sessions
cannot always be reset to a known clean baseline. The same prompt issued
through different accounts may produce variation that cannot be attributed to a
single factor. Privacy policies may describe categories of data collection, but
they rarely specify whether or how those data shape response generation.
Without disclosure, researchers cannot distinguish meaningful personalization
from random variation, safety layer effects, or hidden stratification.

This opacity has equity implications. If models incorporate signals correlated
with race, income, geography, language use, disability, or health literacy,
response variation may reproduce existing disparities. A system might provide
more complete information to some users and more dismissive responses to
others, or it might apply different thresholds for recommending care. Such
differences would not appear as factual errors, but they could shape health
decisions.

% ── Barrier 3 ───────────────────────────────────────────────
\section{Technical Implementation}

The third barrier is technical implementation. Independent evaluation requires
testing systems under conditions that resemble actual use. Large-scale
browser-based testing is restricted by terms of service, rate limits, bot
detection, CAPTCHA challenges, device fingerprinting, and traffic filtering.
Prior work on web scraping notes that terms-of-service violations are generally
treated as civil contract issues, but platforms may respond with account
termination, IP blocking, cease-and-desist letters, or
litigation.~\cite{brown2024scraping} These risks make public-interest research
difficult to conduct openly.

API access is easier to standardise but is not equivalent to browser use. If
personalisation occurs through account memory, consumer interface routing,
cookies, browsing context, or safety layers specific to the product interface,
API-based studies may miss the behaviour of interest. Researchers are placed in
a position comparable to security researchers before safe harbour norms
matured: the work has clear public-interest value, but no reliable permission
pathway exists.

% ── Barrier 4 ───────────────────────────────────────────────
\section{Evaluation Criteria}

The fourth barrier is evaluation criteria. Detecting response differences is
not enough; researchers must determine whether the differences matter.
Accuracy-based evaluation is insufficient because health LLMs can influence
users through tone, framing, omission, and validation. A response may be
factually correct while over-reassuring a user, validating distrust of
clinicians, or discouraging appropriate care.

Manual review can capture subtle forms of sycophancy, but it is difficult to
standardise across reviewers. LLM-as-judge approaches offer scale, but
evaluator models may share alignment tendencies with the systems being
evaluated.~\cite{gu2025llmjudge} A model trained to prefer polished,
agreeable, or user-validating answers may rate those answers favourably.
Evaluation therefore requires patients, clinicians, social scientists, and
ethicists working alongside technical benchmarks.

% ── Barrier 5 ───────────────────────────────────────────────
\section{Temporal Stability}

The fifth barrier is temporal stability. Consumer-facing LLMs change
frequently, often without public changelogs detailed enough for scientific
reproducibility. Model versions, system prompts, routing logic, safety
classifiers, memory features, and personalisation mechanisms may all change
without notice. Some frontier model documentation has raised concerns about
evaluation awareness, which introduces the possibility that model behaviour may
differ under evaluation-like conditions.~\cite{anthropic2025claude}

This would be unacceptable in most clinical research. Trials specify the
intervention under study: formulation, dose, device version, protocol, and
comparator. Consumer LLM studies often cannot specify the equivalent details.
A harmful behaviour documented today may disappear after a silent update; a
safe behaviour observed today may regress tomorrow. Without version
traceability, independent replication becomes difficult and accountability
becomes fragile.

% ── Discussion ──────────────────────────────────────────────
\section{Discussion}

These barriers create a structural governance gap. Claims about safety,
consistency, and fairness cannot be verified if the systems cannot be
independently tested in the settings where patients actually use them. Better
prompts and larger benchmarks will help, but they will not solve the central
problem if researchers remain unable to examine browser-based, personalised,
changing products.

Several reforms are needed. First, developers should disclose which categories
of user signals can influence health-related responses. This does not require
revealing proprietary algorithms, but it should clarify whether location,
cookies, account history, memory, prior conversations, or other contextual
variables affect outputs. Second, consumer-facing systems should expose stable
version identifiers and meaningful changelogs for model, safety, and
personalisation changes that affect health-related guidance. Third, platforms
should create researcher access programmes and safe harbour policies that
permit independent teams to test browser-equivalent workflows at scale under
privacy safeguards, rate limits, audit logs, and responsible disclosure
procedures.

Finally, health-related LLMs should be subject to post-deployment monitoring.
Medical device regulation offers an imperfect but instructive analogy. Under
21 CFR Part 822, the US Food and Drug Administration can require postmarket
surveillance for certain marketed devices through systematic collection and
analysis of safety and effectiveness data.~\cite{cfr822} The EU medical device
framework similarly requires postmarket surveillance within quality management
systems.~\cite{mdr2017} LLMs used for patient-facing guidance should face
comparable expectations: ongoing monitoring, periodic safety reporting, and
structured collection of real-world performance data.

Consumer-facing LLMs already function as informal entry points into health
information and decision-making. Evidence that sycophantic responses can alter
judgment, increase trust, and promote continued use makes independent
evaluation urgent. The question is whether researchers, clinicians, and
regulators will be given the infrastructure needed to study the systems
patients are already using.

% ── Figure ──────────────────────────────────────────────────

% ============================================================
%  REFERENCES
% ============================================================
\bibliographystyle{unsrtnat}
\bibliography{references}

% ============================================================
%  ACKNOWLEDGEMENTS
% ============================================================
\section*{Acknowledgements}

RG reports funding from the Johns Hopkins Institute for Clinical and
Translational Research (ICTR) and Grant Number T32TR004928 from the National
Center for Advancing Translational Sciences (NCATS), a component of the
National Institutes of Health (NIH) and NIH Roadmap for Medical Research. Its
contents are solely the responsibility of the authors and do not necessarily
represent the official view of the Johns Hopkins ICTR, NCATS or NIH.

NJ reports funding from Canadian Institutes of Health Research,
the Fonds de recherche du Qu\'ebec in partnership with Unit\'e de soutien SSA
Qu\'ebec, and the Research Institute of the McGill University Health Centre,
outside the submitted work.

LAC reports grant support from the NIH through DS-I Africa U54
TW012043-01 and Bridge2AI OT2OD032701, the National Science Foundation through
ITEST~\#2148451, and a grant of the Korea Health Technology R\&D Project
through the Korea Health Industry Development Institute (KHIDI), funded by the
Ministry of Health \& Welfare, Republic of Korea (grant number
RS-2024-00403047).

The other authors have nothing to disclose.

\end{document}